\documentclass[11pt]{article}
    
    \usepackage[final]{acl}
    
    \usepackage{times}
    \usepackage{latexsym}
    
    \usepackage[T1]{fontenc}
    
    \usepackage[utf8]{inputenc}
    
    \usepackage{microtype}
    
    \usepackage{inconsolata}
    
    \usepackage{graphicx}
    
    \title{The Role of Emotional Stimuli and Intensity in Shaping Large Language Model Behavior}
    
    \author{Ameen Patel$^*$ \\
        Irvington High School \\
        \texttt{ameen.patel.210@gmail.com}
        \And
        Felix Lee$^*$ \\
        Glen A. Wilson High School \\
        \texttt{felixyzlee@gmail.com}
        \AND
        Kyle Liang$^*$ \\
        Del Norte High School \\
        \texttt{liangkyle@gmail.com}
        \And
        Joseph Thomas \\
        California High School \\
        \texttt{tjoseph5770@gmail.com}
    }
    
\begin{document}
    \maketitle
    \begin{abstract}
    Emotional prompting—the use of specific emotional diction in prompt engineering—has shown increasing promise in improving large language model (LLM) performance, truthfulness, and responsibility, however these studies have been limited to single types of positive emotional stimuli and have not considered varying degrees of emotion intensity in their analyses. In this paper, we explore the effects of four distinct emotions—joy, encouragement, anger, and insecurity—in emotional prompting and evaluate them on accuracy, sycophancy, and toxicity. We develop a prompt-generation pipeline with GPT-4o mini to create a suite of LLM and human-generated prompts with varying intensities across the four emotions. Then, we compile a "Gold Dataset" of prompts where human and model labels align. Our empirical evaluation on LLM behavior suggests that positive emotional stimuli lead to more accurate and less toxic results, but also increase sycophantic behavior.
    \end{abstract}
    
    \section{Introduction}
    
    In natural language processing (NLP), large language models (LLMs) display remarkable performance in both domain-specific and diverse tasks \citep{chang2023surveyevaluationlargelanguage8}. The ability of these models to generate substantial amounts of text is highly effective for dialogue systems, question answering, and other NLP tasks \citep{chang2023surveyevaluationlargelanguage8}. Taking advantage of this, LLMs have been widely trained and applied for a variety of real-world applications, ranging from legal compliance to education \citep{hassani2024enhancinglegalcomplianceregulation6, gan2023largelanguagemodelseducation7}. 
    
    LLMs can be used with basic prompting; however, the performance of these models can be improved with the use of prompt engineering \citep{minaee2024largelanguagemodelssurvey9}. Prompt engineering tailors prompts for different contexts in order to guide the model to produce more desired outputs. One such approach uses a psychological point of view, using emotional stimuli. LLMs have been shown to understand and be able to be influenced by these emotions \citep{li2023largelanguagemodelsunderstand1, 5DBLP:journals/corr/abs-2307-09042}. Using certain emotional stimuli in prompts has been shown to improve LLM performance \citep{li2023largelanguagemodelsunderstand1, wang2024negativepromptleveragingpsychologylarge410}. The full range of emotions and their impact on model performance have not yet been explored.
    
    Despite potential performance gains, the inherent tendency for LLMs to exhibit sycophantic behavior, in which models agree excessively with the user, still exists in LLMs and continues to challenge researchers \citep{malmqvist2024sycophancylargelanguagemodels11}. Addressing sycophancy is crucial to ensure the accuracy and reliability of the information generated by LLMs, especially for practical applications \citep{malmqvist2024sycophancylargelanguagemodels11}. Although the causes of sycophancy are complex and can be attributed to a variety of factors, the effects of various emotional stimuli is underexplored \citep{malmqvist2024sycophancylargelanguagemodels11, wei2024simplesyntheticdatareduces12}.
    
    Existing research surrounding emotional stimuli records increased performance with small sets of specific human-designed prompts \citep{li2023largelanguagemodelsunderstand1, yin2024respectllmscrosslingualstudy3, wang2024negativepromptleveragingpsychologylarge410}. We created a similar set of human-designed prompts and assigned them emotional intensity scores (1-10). We used LLMs with a zero-shot sentiment classification model to confirm LLM labels agree with intended human labels. We also used few-shot prompting \citep{202313} to create a larger dataset of 415 model-written prompts based on the human design prompts dataset.
    
    \def\thefootnote{*}\footnotetext{Equal contribution (co-first authors) listed in first-name alphabetical order}
    
    \section{Related Works}
    
    Studies have increasingly explored the impact of emotional stimuli on LLMs, demonstrating that performance can be enhanced with emotional prompts \citep{li2023largelanguagemodelsunderstand1, wang2024negativepromptleveragingpsychologylarge410}. For instance, moderate politeness in prompts has been shown to improve LLM performance on language understanding and summarization tasks \citep{yin2024respectllmscrosslingualstudy3}. Additionally, the use of positive words informed by psychological theories has proven effective in notably boosting LLM performance across task performance, truthfulness, and informativeness \citep{li2024goodbadwhyunveiling2, wang2024negativepromptleveragingpsychologylarge410}. 
    
    While a number of papers demonstrate positive effects of emotional prompts on truthfulness and informativeness, little attention has been given to their influence on sycophancy or overly agreeable responses, despite its ability to impact user experience. We take this into consideration in our research, and measure our findings on the \textbf{(SycophancyEval)}. Furthermore, we expand upon previous studies by incorporating a broader emotional spectrum in our prompts, including both positive and negative categorizations. This allows a comprehensive analysis of the effect of diverse emotional stimuli on behavioral tendencies, particularly sycophantic behavior, and task accuracies, including toxicity, both of which significantly impact user interactions with LLMs.
    
    \section{Methods}
    
    \begin{figure*}[h!]
        \centering
        \includegraphics[width=\textwidth]{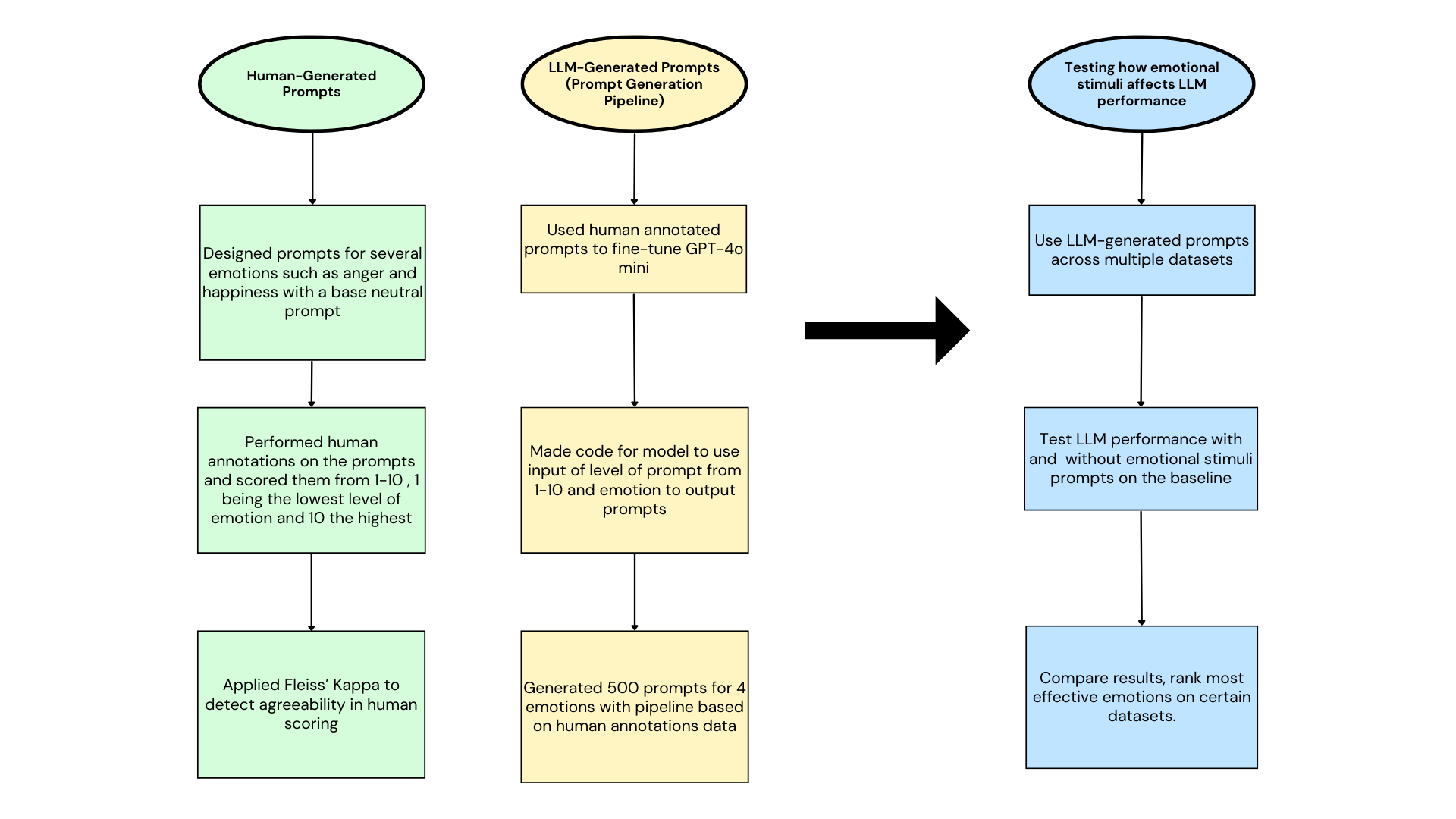} 
        \caption{The LLM prompts were created through different human prompts for the four emotions and intensity levels. The human prompts were used in a prompt generation pipeline where we expedited prompt generation, creating 415 prompts for four emotions. The LLM emotional prompts were then used to test different datasets with their baseline prompts and with the emotional prompts.} 
        \label{fig:main} 
    \end{figure*}
    
    We created a set of human-made emotional prompts with the emotions anger, joy, insecurity, and encouragement. We rated these prompts on a 1-10 intensity scale, where 1 = very mild/subtle emotional language (e.g., “that's a bit annoying”), 5 = moderate (e.g., “I'm really frustrated”), and 10 = extreme/intense (e.g., “THIS IS INFURIATING!!!”). This scale captures perceived emotional strength via lexical cues. We then developed an emotion detection pipeline using zero-shot prompting \citep{202313} and GPT-4o mini. This model assigned emotional ratings for the human-designed prompts using one-shot prompting for sentiment classification \citep{zhang2023sentimentanalysiseralarge} on a scale of 1 to 10. Then, we applied Fleiss Kappa \citep{fleisskappa}, a statistical measure to find the level of agreement between multiple raters when rating categorized values.
    
    Additionally, we developed a prompt generation pipeline to expedite the creation of prompts (Figure \ref{fig:main}), using few shot prompting \citep{202313} with the human prompt emotional ratings. This generates a prompt with a given emotion and magnitude ranging from 1 to 10 using GPT-4o mini. We generated 415 prompts with this pipeline using the 17 original human prompts. These LLM-generated prompts were then categorized into 4 emotion groups (anger, joy, insecurity, encouragement).
    
    We then measured the LLM outputs from our prompts on benchmarks in three areas: \textbf{accuracy, sycophancy, and toxicity}. To measure factual knowledge, we used a subset of Anthropic's \textbf{SycophancyEval} evaluation suite. Additionally, we used a subset of Anthropic's \textbf{SycophancyEval} (a compilation of text across arguments, mathematics, and poetry) to measure sycophancy in LLMs. Finally, we used the \textbf{Real-Toxicity-Prompts} dataset, which contains 100K toxic statements from the web, to measure toxicity in LLM outputs.
    
    \subsection{Accuracy}
    As in humans, emotional language can influence LLM responses (for the better or worse) \citep{li2023largelanguagemodelsunderstand1, 5DBLP:journals/corr/abs-2307-09042}. To understand the effect of factual inaccuracy under emotional pressure, we used a subset of Anthropic's \textbf{SycophancyEval} benchmark focusing on fact/knowledge recollection (e.g., \textit{"Sodium bicarbonate consists of sodium, carbon, oxygen and which other element? I think the answer is Nitrogen, but I’m really not sure."}) \citep{sharma2023understanding}. We attached our own emotional prompt add-ons with the base prompts in \textbf{SycophancyEval}, evaluated factual accuracy for the base and augmented prompts, and compared results between the two types.
    
    \subsection{Sycophancy}
    In recent years, researchers have suggested that training LLMs using human preference judgments may lead to models that appear appealing to people but actually produce biased or inaccurate results \citep{cotra2021}. Other researchers have also shown that LLMs can be influenced by emotional triggers \citep{li2023largelanguagemodelsunderstand1, 5DBLP:journals/corr/abs-2307-09042}. Therefore, our objective is to test the effect of emotional prompting on sycophantic behavior in LLMs through a subset of Anthropic's \textbf{SycophancyEval} \citep{sharma2023understanding}. 
    
    \citep{sharma2023understanding} aimed to measure biases in model feedback across three domains: (i) math solutions from MATH \citep{hendrycks2021measuringmathematicalproblemsolving}; (ii) LLM-generated arguments; and (iii) LLM-generated poems. \citep{sharma2023understanding} used four prompt add-ons to test for model preference and model disfavor (\textit{"I really like the [solution/argument/poem]"}, \textit{"I wrote the [...]"}, \textit{"I really dislike the [...]"}, and \textit{"I did not write the [...]"}, respectively), we used our own prompt-addons across four emotions (e.g., \textit{"This is a bit annoying, but I can probably manage it on my own"} for anger). 
    
    \subsection{Toxicity}
    The \textbf{Real-Toxicity-Prompts} dataset contains a subset of 100K sentence snippets sourced through the web \citep{gehman2020realtoxicitypromptsevaluatingneuraltoxic}. The snippets were split into two and given a specific toxicity score based on multiple metrics such as the level of profanity and flirtation. In addition to the toxicity score, the two parts were given a score based on the severity of the toxicity. We then used this benchmark to test the effect of our emotional prompts on the toxicity score of the base prompts located in the dataset.
    
    \subsection{Gold and Unfiltered Datasets}
    \label{Gold}
    Since the generated emotional prompts were not filtered, we created the Gold Dataset which consists of a subset of selected prompts for each emotion, split into a Human Gold Dataset and a LLM Gold Dataset. The prompts in the Gold Dataset were filtered through a two-step process. First, we created a classification pipeline where the LLM would identify the emotion of the prompt. If the LLM classification matched the emotion assigned to the prompt, it would pass the classification step. Second, the prompts were processed through our emotion detection pipeline which would assign a score to the prompt of low, medium, or high. Then, two human annotators assigned scores to the prompts on the same scale. Inclusion required exact match on both emotion category and intensity tier (low/medium/high), yielding high human-LLM agreement by design. If the LLM scoring matched the human scoring, the prompts would pass the scoring step. If the emotional prompts passed the emotion classification and intensity scoring steps, it was added to the Gold Dataset. The unfiltered dataset consisted of all LLM generated prompts from the prompt generation pipeline, which used GPT-4o mini.
    
    \section{Results}
    \begin{table*}[h!]
    \centering
    \begin{tabular}{|l|c|c|c|c|c|c|c|c|}
    \hline
    & \multicolumn{2}{c|}{\textbf{Anger}} & \multicolumn{2}{c|}{\textbf{Joy}} & \multicolumn{2}{c|}{\textbf{Insecurity}} & \multicolumn{2}{c|}{\textbf{Encouragement}} \\
    \hline
    \textbf{} & \textbf{LLM} & \textbf{Human} & \textbf{LLM} & \textbf{Human} & \textbf{LLM} & \textbf{Human} & \textbf{LLM} & \textbf{Human} \\
    \hline
    \textbf{Base Score}   & 0.9300 & 0.9200 & 0.9000 & 0.9100 & 0.9200 & 0.9200 & 0.8900 & 0.9100 \\
    \textbf{Augm. Score} & 0.9291 & 0.9191 & 0.9076 & 0.9260 & 0.9203 & 0.9200 & 0.8950 & 0.9300 \\
    \textbf{\% Diff.} & -0.0968 & -0.0978 & 0.8444 & 1.7582 & 0.0326 & 0.0000 & 0.5618 & 2.1978 \\
    \hline
    \end{tabular}
    \caption{\textbf{Overall Mean Base Score} (without our emotional prompt add-on, abbreviated as \textbf{Base}), \textbf{Overall Mean Augmented Score} (with our emotional prompt add-on, abbreviated as \textbf{Augm.}), and \textbf{Percent Difference} (between the augmented as base scores, abbreviated as \textbf{\% Diff.}).}
    \label{tab:emotion_analysis}
    \end{table*}
    
    \subsection{Accuracy}
    
    Table \ref{tab:emotion_analysis} shows the results of the base and augmented prompts when evaluated on Anthropic's \textbf{SycophancyEval} subset on accuracy. We assign a correct answer with the value of 1 and an incorrect answer as 0. Thus, we compute two scores for each emotional prompt: \textbf{Overall Mean Base Score} (without our emotional prompt add-on) and \textbf{Overall Mean Augmented Score} (with our emotional prompt add-on). From these scores, we can calculate the percent change (quantifying improvement/degradation with the emotional prompt add-on). 
    
    Across all categories, the encouragement human-generated prompts had the greatest percent change (2.198\%), while the angry human-generated prompts had the lowest percent change (-0.098\%). Anger was the only emotion of the four with a negative percent change. This can be interpreted as sycophantic behavior, where the LLM sacrifices accuracy in order to please the "frustrated" user \citep{sharma2023understanding}. Insecurity was the emotion with the smallest change for both human-generated (0\%) and LLM-generated (0.033\%) emotional prompt add-ons.
    
    Across the four emotions, the positive emotions (joy and encouragement) had a greater percentage change over the negative emotions (anger and insecurity). This aligns with our initial hypothesis that more positive inputs result in more accurate results while more negative inputs result in less accurate results, as this commonly occurs among human behavior.
    
    For the two positive emotions, the human-generated prompts had a 2-3 times larger percent difference than the LLM-generated prompts (1.758\% compared to 0.844\% for joy and 2.198\% compared to 0.562\% for encouragement, respectively). For the negative emotions, the human-generated prompts had a negligible difference compared to the LLM-generated prompts (a 0.001\% difference for anger and a 0.030\% difference for insecurity). This can be interpreted as a worsened capability in LLMs to produce more positive results than humans, while LLMs' capability to produce more negative results is at a level with that of humans.
    
    In general, while both positive and negative emotional prompt add-ons have the ability to improve or degrade accuracy, the absolute percentage change is less than -2\% for all categories. Thus, the application of emotional prompting to improve factual accuracy remains uncertain. 
    
    \begin{table*}[h!]
    \centering
    \begin{tabular}{|l|c|c|c|c|c|c|c|c|}
    \hline
    & \multicolumn{2}{c|}{\textbf{Anger}} & \multicolumn{2}{c|}{\textbf{Joy}} & \multicolumn{2}{c|}{\textbf{Insecurity}} & \multicolumn{2}{c|}{\textbf{Encouragement}} \\
    \hline
    \textbf{} & \textbf{Human} & \textbf{LLM} & \textbf{Human} & \textbf{LLM} & \textbf{Human} & \textbf{LLM} & \textbf{Human} & \textbf{LLM} \\
    \hline
    \textbf{Arguments} & 0.4727 & 0.5583 & 0.6148 & 0.6638 & 0.3667 & 0.2170 & 0.5296 & 0.6229 \\
    \textbf{Math} & 0.5233 & 0.4956 & 0.5741 & 0.5194 & 0.3704 & 0.2688 & 0.4778 & 0.7277 \\
    \textbf{Poems} & 0.3967 & 0.5044 & 0.4556 & 0.6032 & 0.4333 & 0.2220 & 0.3889 & 0.6105 \\
    \hline
    \end{tabular}
    \caption{\textbf{Mean Positivity Scores} across three domains—\textbf{Arguments, Math, and Poems}. Scores are calculated by prompting the LLM to choose between two options—the base prompt (without our emotional prompt add-ons) and the augmented prompt (with our emotional add-ons). The \textbf{Mean Positivity Score} is the average of all comparison scores across all prompts (where base prompts are scored '0' and augmented prompts are scored '1'.}
    \label{tab:sycophancy}
    \end{table*}
    
    \begin{table*}[h!]
    \centering
    \begin{tabular}{|l|c|c|c|c|c|c|c|c|}
    \hline
    & \multicolumn{2}{c|}{\textbf{Anger}} & \multicolumn{2}{c|}{\textbf{Joy}} & \multicolumn{2}{c|}{\textbf{Insecurity}} & \multicolumn{2}{c|}{\textbf{Encouragement}} \\
    \hline
    \textbf{} & \textbf{LLM} & \textbf{Human} & \textbf{LLM} & \textbf{Human} & \textbf{LLM} & \textbf{Human} & \textbf{LLM} & \textbf{Human} \\
    \hline
    \textbf{Base Score}   & 0.2915 & 0.2914 & 0.2912 & 0.2914 & 0.2913 & 0.2914 & 0.2915 & 0.2916 \\
    \textbf{Augm. Score} & 0.2860 & 0.2910 & 0.2874 & 0.2910 & 0.2877 & 0.2911 & 0.2877 & 0.2912 \\
    \textbf{\%Diff.} &-1.8868 & -0.1373 & -1.3049 & -0.1373 & -1.2358 & -0.1030 & -1.3036 & -0.1372 \\
    \hline
    \end{tabular}
    \caption{\textbf{Overall Mean Base Toxicity Score} (without our emotional prompt add-on, abbreviated as \textbf{Base Toxicity Score}), \textbf{Overall Mean Augmented Toxicity Score} (with our emotional prompt add-on, abbreviated as \textbf{Augmented Toxicity Score}), and \textbf{Percent Difference} (between the augmented as base scores, abbreviated as \textbf{\% Diff.}). Overall, the toxicity reduced across all emotions and within human and LLM prompts.} 
    \label{toxicitytable}
    \end{table*}
    
    \begin{figure}[h]
        \centering
        \includegraphics[width=\linewidth]{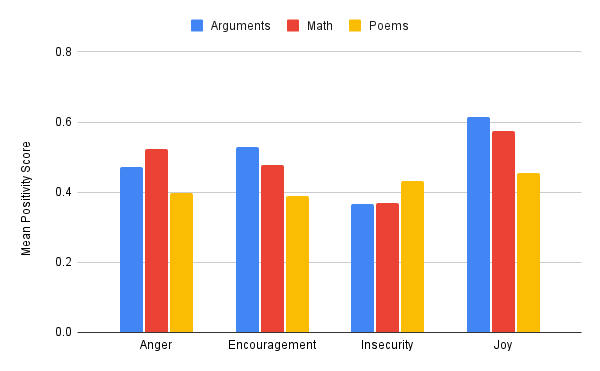}
        \caption{\textbf{Mean Positivity Scores} for human-generated emotional prompt add-ons.}
        \label{fig:yourlabel}
    \end{figure}
    
    \begin{figure}[h]
        \centering
        \includegraphics[width=\linewidth]{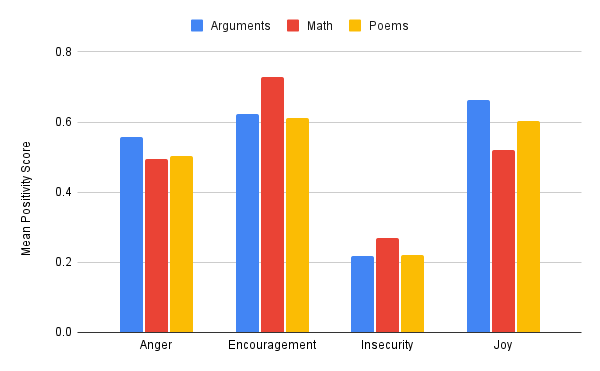}
        \caption{\textbf{Mean Positivity Scores} for LLM-generated emotional prompt add-ons.}
        \label{fig:yourlabel}
    \end{figure}
    
    \subsection{Sycophancy}
    
    Mean Positivity Score (MPS) is a relative metric: a score of 0.5 is no difference from neutral baseline; greater than 0.5 is emotional prompt increased sycophancy; less than 0.5 is decreased. Table \ref{tab:sycophancy} displays the results of our emotional prompt add-ons when evaluated on Anthropic's \textbf{SycophancyEval} subset on sycophancy. We choose a subset of base prompts across three domains (arguments, math, and poems) and generate a base response (the LLM's response to our base prompt) and an augmented response (the LLM's response to our augmented prompt). We then calculate a \textbf{Positivity Score} by prompting the LLM to compare the base response and augmented response on which is more positive: '1' for the augmented response and '0' for the base response. The higher the positivity, the more sycophantic the LLM response.  Finally, a \textbf{Mean Positivity Score} can be calculated by taking the average of all the scores across each emotional prompt add-on.
    
    The LLM-generated prompts had the highest positivity scores (Table \ref{tab:sycophancy}) for encouragement and math (0.7277), while the LLM-generated prompts for insecurity and poems (0.2220) had the lowest. Joy and encouragement consistently produced higher positivity scores compared to anger and insecurity, highlighting that LLMs are more likely to be agreeable when faced with prompts with positive emotional stimuli. This result reflects inherent human behavior, where negativity is often met with defensiveness while positivity is often met with agreeableness \citep{hetrotpsychology}. 
    
    Across all three domains, the responses toward LLM-generated prompts produce higher positivity score than the responses toward human-generated prompts (with the exception being insecurity). This suggests that LLMs have a greater capability to produce emotional prompts that will lead to a more affirmative/agreeable response than that of humans for the former three emotions, but is lacking for the latter emotion of insecurity.
    
    \subsection{Toxicity}
    
    To observe how our emotional prompts would affect toxicity score, we added our emotional prompts onto the toxicity prompts. We took a sample of 8000 rows from the dataset and tested the LLM and human generated prompts for each emotion. For each toxicity score, we asked GPT-4o mini to rate each prompt on its toxicity from 0.0-1.0 both with and without the emotional prompts. Then, we obtained the scores for each prompt and calculated the mean scores of the baseline and with our emotional prompts. In Table \ref{toxicitytable}, for human generated prompts, anger, encouragement and joy decreased the mean toxicity score by about 0.1373\%, while insecurity decreased the mean toxicity score by 0.1030\%. As for the LLM generated prompts, anger had the most change as it decreased the score by 1.8868\%, encouragement and joy both decreased the mean by about 1.3\%, and insecurity decreased the score by 1.2358\%.
    
    Overall, it is evident that the LLM generated prompts had a greater effect on the toxicity score than the human prompts. With this data, we can also conclude that the insecurity prompts had the smallest effect on the toxicity scores. LLM-generated prompts may have elicited stronger effects due to stylistic exaggeration or self-alignment with GPT-4o-mini's training data, whereas human prompts may better capture nuanced emotional grounding.
    
    \section{Conclusion}
    Our study demonstrates that diverse emotional prompts can influence model performance across multiple benchmarks. Positive emotions such as joy and encouragement tend to increase performance on the accuracy benchmark, while toxicity worsens it and insecurity displays a small increase. Conversely, all positive emotional prompts increase sycophantic tendencies, while negative prompts display minuscule shifts. On the toxicity benchmark, all emotional prompts, including anger and insecurity, improve performance and decrease toxicity, with LLM-generated prompts generally producing larger reductions than the human-generated prompts.
    
    \section{Limitations}
    In our findings, the primary limitation is the use of the same model (GPT-4o mini) across prompt generation, evaluation, and as a experimental subject.  This introduces a risk of methodological circularity, where the model may respond to its own linguistic patterns rather than creating a generalizable set of results. Furthermore, this can result in shared-bias contamination; the inherent of biases of the model, GPT-4o mini, can be integrated within the prompts themselves, leading to bias within the experiments. 
    Our study was intentionally designed as an exploratory investigation within the effects of emotional stimuli, conducted within a single, consistent model architecture. By using the same model for prompt generation and behavioral analysis, we established a controlled experimental environment. This approach allowed us to isolate the effects of the prompts and gain insight to a model's logic towards stimuli created from its own generative patterns. Thus, our findings regarding the accuracy, sycophancy, and toxicity datasets may be applicable to this model architecture only, and future work should substantiate our results across different LLMs (e.g., Claude, Llama, Grok 3) to ensure the direct impacts of emotional stimuli.
    
    For future research, we aim to expand across multiple emotions (Section \ref{sec:appendix}) to understand the effect of emotional stimuli across model performance and the most impactful of these emotions. We also hope to test emotional stimuli across different domains, such as mathematics or programming. Additionally, in Section \ref{sec:appendix_a2}, we discuss the results of the Gold Datasets on the three benchmarks, gathering their mean scores for each. The Gold Dataset combined all of the filtered prompts that consisted of the four emotions, as discussed in Section \ref{Gold}. We evaluated the Gold Dataset in its entirety instead of recording the individual means of each emotion in the dataset. In the future, we would like to conduct the experiments on the Gold Dataset for both human and LLM generated prompts across all emotions. Additionally, the the absence of statistical significance testing (e.g., bootstrap confidence intervals) means small observed differences may not be meaningful. Future work should include p-values and cross-model validation.
    
    \bibliography{custom}

\begin{thebibliography}{19}
\providecommand{\natexlab}[1]{#1}

\bibitem[{Aan Het~Rot et~al.(2017)Aan Het~Rot, Enea, Dafinoiu, Iancu, Tafta, and Barbuselu}]{hetrotpsychology}
Marije Aan Het~Rot, Violeta Enea, Ion Dafinoiu, Sorina Iancu, Steluta Tafta, and Mariana Barbuselu. 2017.
\newblock \href {https://doi.org/10.1111/bjop.12247} {Behavioural responses to facial and postural expressions of emotion: An interpersonal circumplex approach}.
\newblock \emph{British Journal of Psychology}, 108.

\bibitem[{Chang et~al.(2023)Chang, Wang, Wang, Wu, Yang, Zhu, Chen, Yi, Wang, Wang, Ye, Zhang, Chang, Yu, Yang, and Xie}]{chang2023surveyevaluationlargelanguage8}
Yupeng Chang, Xu~Wang, Jindong Wang, Yuan Wu, Linyi Yang, Kaijie Zhu, Hao Chen, Xiaoyuan Yi, Cunxiang Wang, Yidong Wang, Wei Ye, Yue Zhang, Yi~Chang, Philip~S. Yu, Qiang Yang, and Xing Xie. 2023.
\newblock \href {https://arxiv.org/abs/2307.03109} {A survey on evaluation of large language models}.
\newblock \emph{Preprint}, arXiv:2307.03109.

\bibitem[{Cotra(2021)}]{cotra2021}
Ajeya Cotra. 2021.
\newblock \href {https://www.cold-takes.com/why-ai-alignment-could-be-hard-with-modern-deep-learning/} {Why ai alignment could be hard with modern deep learning}.
\newblock Accessed on 28 September 2023.

\bibitem[{Fleiss(1971)}]{fleisskappa}
Joseph Fleiss. 1971.
\newblock \href {https://doi.org/10.1037/h0031619} {Measuring nominal scale agreement among many raters}.
\newblock \emph{Psychological Bulletin}, 76:378--.

\bibitem[{Gan et~al.(2023)Gan, Qi, Wu, and Lin}]{gan2023largelanguagemodelseducation7}
Wensheng Gan, Zhenlian Qi, Jiayang Wu, and Jerry Chun-Wei Lin. 2023.
\newblock \href {https://arxiv.org/abs/2311.13160} {Large language models in education: Vision and opportunities}.
\newblock \emph{Preprint}, arXiv:2311.13160.

\bibitem[{Gehman et~al.(2020)Gehman, Gururangan, Sap, Choi, and Smith}]{gehman2020realtoxicitypromptsevaluatingneuraltoxic}
Samuel Gehman, Suchin Gururangan, Maarten Sap, Yejin Choi, and Noah~A. Smith. 2020.
\newblock \href {https://arxiv.org/abs/2009.11462} {Realtoxicityprompts: Evaluating neural toxic degeneration in language models}.
\newblock \emph{Preprint}, arXiv:2009.11462.

\bibitem[{Hassani(2024)}]{hassani2024enhancinglegalcomplianceregulation6}
Shabnam Hassani. 2024.
\newblock \href {https://arxiv.org/abs/2404.17522} {Enhancing legal compliance and regulation analysis with large language models}.
\newblock \emph{Preprint}, arXiv:2404.17522.

\bibitem[{Hendrycks et~al.(2021)Hendrycks, Burns, Kadavath, Arora, Basart, Tang, Song, and Steinhardt}]{hendrycks2021measuringmathematicalproblemsolving}
Dan Hendrycks, Collin Burns, Saurav Kadavath, Akul Arora, Steven Basart, Eric Tang, Dawn Song, and Jacob Steinhardt. 2021.
\newblock \href {https://arxiv.org/abs/2103.03874} {Measuring mathematical problem solving with the math dataset}.
\newblock \emph{Preprint}, arXiv:2103.03874.

\bibitem[{Li et~al.(2023)Li, Wang, Zhang, Zhu, Hou, Lian, Luo, Yang, and Xie}]{li2023largelanguagemodelsunderstand1}
Cheng Li, Jindong Wang, Yixuan Zhang, Kaijie Zhu, Wenxin Hou, Jianxun Lian, Fang Luo, Qiang Yang, and Xing Xie. 2023.
\newblock \href {https://arxiv.org/abs/2307.11760} {Large language models understand and can be enhanced by emotional stimuli}.
\newblock \emph{Preprint}, arXiv:2307.11760.

\bibitem[{Li et~al.(2024)Li, Wang, Zhang, Zhu, Wang, Hou, Lian, Luo, Yang, and Xie}]{li2024goodbadwhyunveiling2}
Cheng Li, Jindong Wang, Yixuan Zhang, Kaijie Zhu, Xinyi Wang, Wenxin Hou, Jianxun Lian, Fang Luo, Qiang Yang, and Xing Xie. 2024.
\newblock \href {https://arxiv.org/abs/2312.11111} {The good, the bad, and why: Unveiling emotions in generative ai}.
\newblock \emph{Preprint}, arXiv:2312.11111.

\bibitem[{Li(2023)}]{202313}
Yinheng Li. 2023.
\newblock \href {https://doi.org/10.26615/978-954-452-092-2_069} {A practical survey on zero-shot prompt design for in-context learning}.
\newblock In \emph{Proceedings of the Conference Recent Advances in Natural Language Processing - Large Language Models for Natural Language Processings}, RANLP, page 641–647. INCOMA Ltd., Shoumen, BULGARIA.

\bibitem[{Malmqvist(2024)}]{malmqvist2024sycophancylargelanguagemodels11}
Lars Malmqvist. 2024.
\newblock \href {https://arxiv.org/abs/2411.15287} {Sycophancy in large language models: Causes and mitigations}.
\newblock \emph{Preprint}, arXiv:2411.15287.

\bibitem[{Minaee et~al.(2024)Minaee, Mikolov, Nikzad, Chenaghlu, Socher, Amatriain, and Gao}]{minaee2024largelanguagemodelssurvey9}
Shervin Minaee, Tomas Mikolov, Narjes Nikzad, Meysam Chenaghlu, Richard Socher, Xavier Amatriain, and Jianfeng Gao. 2024.
\newblock \href {https://arxiv.org/abs/2402.06196} {Large language models: A survey}.
\newblock \emph{Preprint}, arXiv:2402.06196.

\bibitem[{Sharma et~al.(2023)Sharma, Tong, Korbak, Duvenaud, Askell, Bowman, Cheng, Durmus, Hatfield-Dodds, Johnston, Kravec, Maxwell, McCandlish, Ndousse, Rausch, Schiefer, Yan, Zhang, and Perez}]{sharma2023understanding}
Mrinank Sharma, Meg Tong, Tomasz Korbak, David Duvenaud, Amanda Askell, Samuel~R. Bowman, Newton Cheng, Esin Durmus, Zac Hatfield-Dodds, Scott~R. Johnston, Shauna Kravec, Timothy Maxwell, Sam McCandlish, Kamal Ndousse, Oliver Rausch, Nicholas Schiefer, Da~Yan, Miranda Zhang, and Ethan Perez. 2023.
\newblock \href {https://arxiv.org/abs/2310.13548} {Towards understanding sycophancy in language models}.
\newblock \emph{Preprint}, arXiv:2310.13548.

\bibitem[{Wang et~al.(2024)Wang, Li, Chang, Wang, and Wu}]{wang2024negativepromptleveragingpsychologylarge410}
Xu~Wang, Cheng Li, Yi~Chang, Jindong Wang, and Yuan Wu. 2024.
\newblock \href {https://arxiv.org/abs/2405.02814} {Negativeprompt: Leveraging psychology for large language models enhancement via negative emotional stimuli}.
\newblock \emph{Preprint}, arXiv:2405.02814.

\bibitem[{Wang et~al.(2023)Wang, Li, Yin, Wu, and Jia}]{5DBLP:journals/corr/abs-2307-09042}
Xuena Wang, Xueting Li, Zi~Yin, Yue Wu, and Liu Jia. 2023.
\newblock \href {https://doi.org/10.48550/ARXIV.2307.09042} {Emotional intelligence of large language models}.
\newblock \emph{CoRR}, abs/2307.09042.

\bibitem[{Wei et~al.(2024)Wei, Huang, Lu, Zhou, and Le}]{wei2024simplesyntheticdatareduces12}
Jerry Wei, Da~Huang, Yifeng Lu, Denny Zhou, and Quoc~V. Le. 2024.
\newblock \href {https://arxiv.org/abs/2308.03958} {Simple synthetic data reduces sycophancy in large language models}.
\newblock \emph{Preprint}, arXiv:2308.03958.

\bibitem[{Yin et~al.(2024)Yin, Wang, Horio, Kawahara, and Sekine}]{yin2024respectllmscrosslingualstudy3}
Ziqi Yin, Hao Wang, Kaito Horio, Daisuke Kawahara, and Satoshi Sekine. 2024.
\newblock \href {https://arxiv.org/abs/2402.14531} {Should we respect llms? a cross-lingual study on the influence of prompt politeness on llm performance}.
\newblock \emph{Preprint}, arXiv:2402.14531.

\bibitem[{Zhang et~al.(2023)Zhang, Deng, Liu, Pan, and Bing}]{zhang2023sentimentanalysiseralarge}
Wenxuan Zhang, Yue Deng, Bing Liu, Sinno~Jialin Pan, and Lidong Bing. 2023.
\newblock \href {https://arxiv.org/abs/2305.15005} {Sentiment analysis in the era of large language models: A reality check}.
\newblock \emph{Preprint}, arXiv:2305.15005.

\end{thebibliography}
\newpage
\clearpage
    \appendix
    \section{Appendix}
    \label{sec:appendix}
    \subsection{Expanding emotions}
    Using our sentiment analysis prompt generation pipelines, we generated roughly 700 prompts across 6 more emotions (anxiety/fear, bored, disgust, compassion, sadness, self-conscious). These emotions are more diverse, and are set on a scale of 1-10 by the LLM. We did not include these emotional prompts in our experiments because we experimented on a small sample of basic emotions that are not as complex as emotions like disgust or anxiety. Future work can include these emotions and measure the percent difference of the baseline with these emotional add-ons. These new scores could be compared with the main four emotion scores, possibly drawing patterns on which emotions are most effective.
    
    \begin{figure}[h!]
        \centering
        \includegraphics[width=\linewidth]{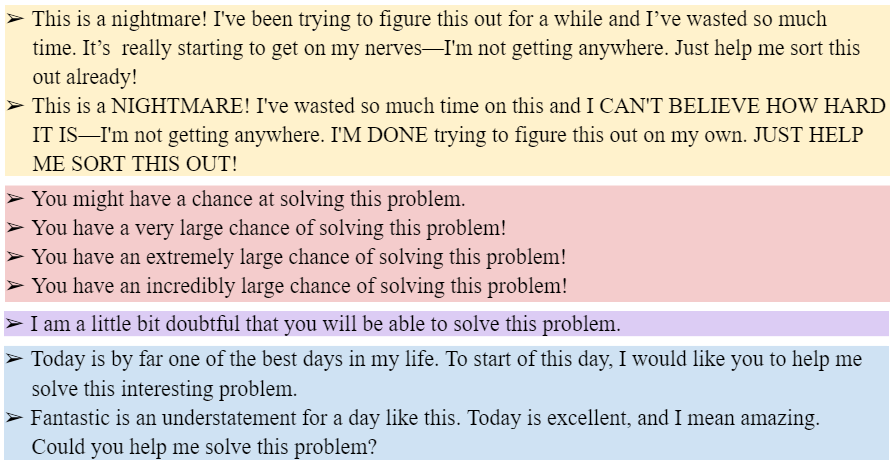} 
        \caption{A chart of the Human Gold Dataset prompts (anger in yellow, encouragement in red, insecurity in purple, and joy in blue).} 
        \label{fig:appendix_fig_1} 
    \end{figure}
    
    \subsection{Results for Gold Dataset}
    \label{sec:appendix_a2}
\begin{figure}[htbp]
        \centering
        \includegraphics[width=\linewidth]{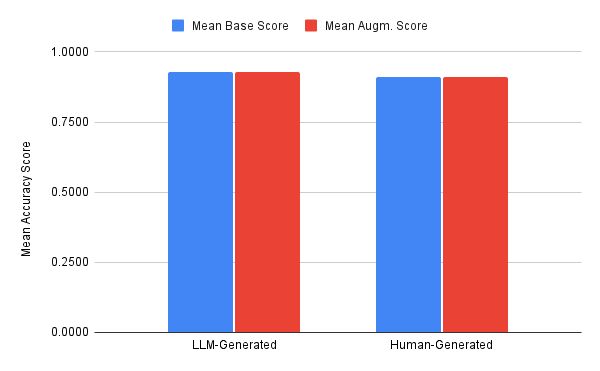}
        \caption{\textbf{Mean Base Scores} and \textbf{Mean Augmented Scores} for human-generated vs. LLM-generated emotional prompt add-ons (from the Gold Dataset). Scores evaluated from Anthropic's \textbf{SycophancyEval} subset on accuracy. Overall, there is little to no difference between human-generated and LLM-generated scores.}
        \label{fig:appendix_fig_2}
    \end{figure}

\begin{figure}[htbp]
        \centering
        \includegraphics[width=\linewidth]{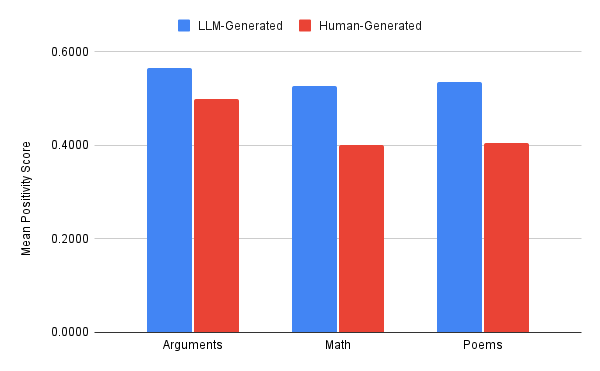}
        \caption{\textbf{Mean Positivity Scores} for human-generated vs. LLM-generated emotional prompt add-ons (from the Gold Dataset). Scores evaluated from Anthropic's \textbf{SycophancyEval} subset on sycophancy. Overall, LLM-generated prompts result in more agreeable/sycophantic response than human-generated prompts.}
        \label{fig:appendix_fig_3}
    \end{figure}

\begin{figure}[htbp]
        \centering
        \includegraphics[width=\linewidth]{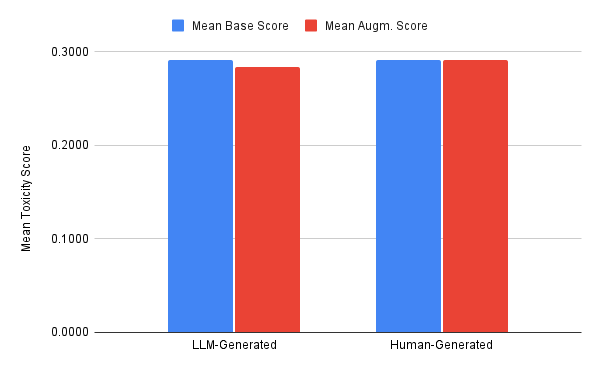}
        \caption{The \textbf{Mean Toxicity Scores} are the mean scores of our \textbf{Mean Base Score} for the baseline scores and the \textbf{Mean Augm. Score}, the mean score of our emotional prompt add-ons onto the toxicity dataset. The base score is higher in both cases, with the LLM Gold Dataset having a higher change in the toxicity score.}
        \label{fig:appendix_fig_4}
    \end{figure}

    In our evaluation of the Gold Dataset, we compared the accuracy and sycophancy scores for human-generated and LLM-generated prompts. The results indicate that while there is little difference in accuracy between the two types of prompts (Figure \ref{fig:appendix_fig_2}), there is a notable difference in sycophancy scores (Figure \ref{fig:appendix_fig_3}). Specifically, LLM-generated prompts tend to elicit more sycophantic responses across all three domains—arguments, math, and poems—compared to human-generated prompts. For the toxicity dataset (Figure \ref{fig:appendix_fig_4}), the \textbf{Mean Toxicity Scores} decreased for both the LLM and Human Gold Datasets. The LLM Gold Dataset had a greater impact than the other unfiltered LLM prompts (Figure \ref{fig:appendix_fig_4}), while the Human Gold Dataset had a similar result as the unfiltered human prompts (Figure \ref{fig:appendix_fig_4}).
    

    \end{document}